# Face-MagNet: Magnifying Feature Maps to Detect Small Faces

Pouya Samangouei* Mahyar Najibi*
Rama Chellappa Larry S. Davis
University of Maryland Institute for Advanced Computer Studies
Baltimore Ave, College Park, MD, 20742
pouya,lsd,rama@umiacs.umd.edu najibi@cs.umd.edu

## Abstract

*In this paper, we introduce the Face Magnifier Network (Face-MageNet), a face detector based on the Faster-RCNN framework which enables the flow of discriminative information of small scale faces to the classifier without any skip or residual connections. To achieve this, Face-MagNet deploys a set of ConvTranspose, also known as deconvolution, layers in the Region Proposal Network (RPN) and another set before the Region of Interest (RoI) pooling layer to facilitate detection of finer faces. In addition, we also design, train, and evaluate three other well-tuned architectures that represent the conventional solutions to the scale problem: context pooling, skip connections, and scale partitioning. Each of these three networks achieves comparable results to the state-of-the-art face detectors. With extensive experiments, we show that Face-MagNet based on a VGG16 architecture achieves better results than the recently proposed ResNet101-based HR [7] method on the task of face detection on WIDER dataset and also achieves similar results on the hard set as our other recently proposed method SSH [17].*[1]

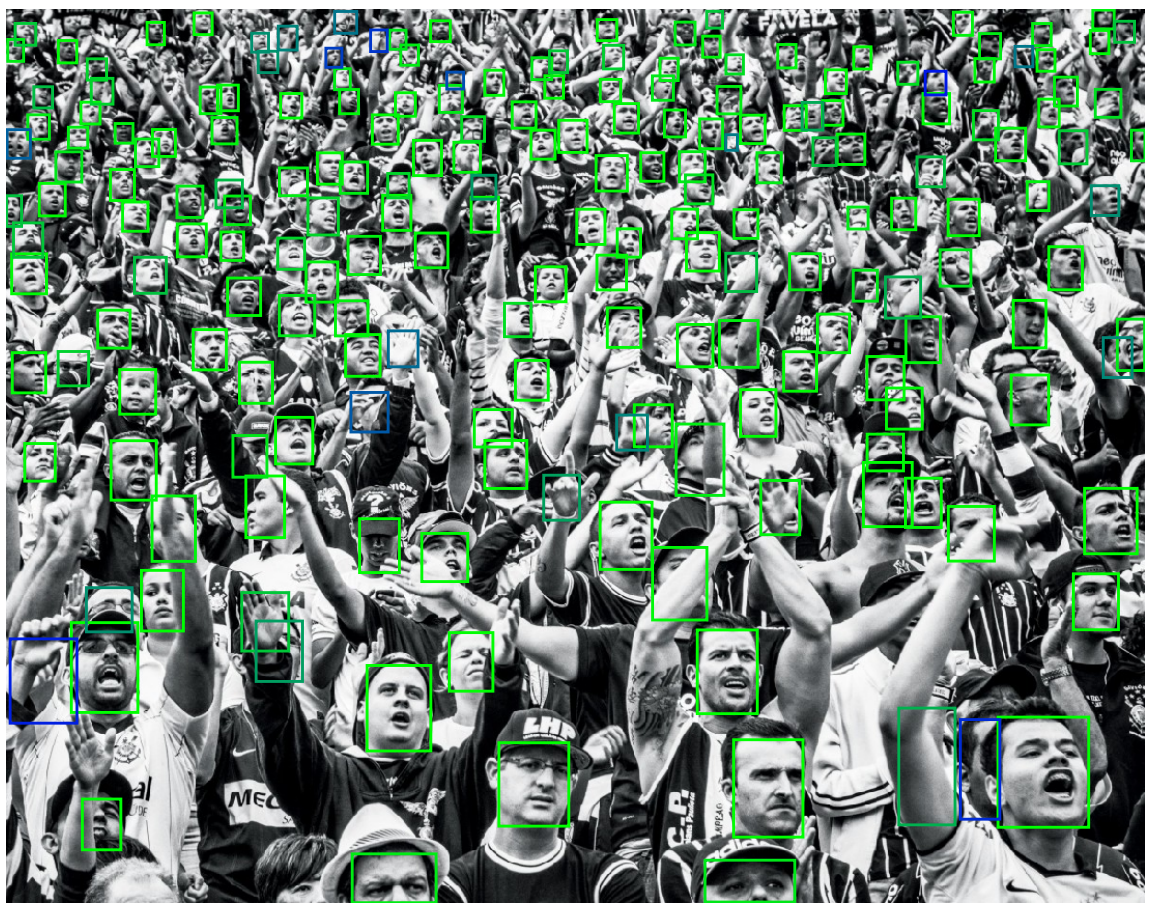

Figure 1: An example of face detection result using the proposed algorithm. The green boxes are the confident detection results and the blue ones are just above the decision boundary.

## 1. Introduction

Object detection has always been an active area of research in computer vision where researchers compare their methods on established benchmarks such as MS-COCO [13]. The main challenge in these benchmarks is the unconstrained environment where the objects appear in. Although, the number of small objects in these benchmarks has been increased recently, it is not enough to affect the final aggregated detection results. Therefore, most of the focus is on challenges rather than the scale of the objects. Besides the WIDER [25] dataset, other face detection benchmarks such as FDDB [8], and Pascal Faces [3] were of no exception regarding this matter. However, as shown in [25] and Figure 2, even the well performing object/face detection methods fail badly on WIDER dataset. We investigate different aspects of detecting small faces on this dataset.

One of the earliest solutions to the scale problem is ”skip connections” [6, 1]. Skip connections resize, normalize, scale and concatenate feature maps with possibly different resolutions to transfer the feature information from earlier convolutional layers to the final classifier. They are already employed for face detection in *HyperFace* [20] and *CMS-RCNN* [27]. We train our variant of this solution as the SkipFace network. Compared to *CMS-RCNN*, our normalization of concatenated layers is much simpler. This network achieves a better result than *CMS-RCNN* and *HR-VGG16* [7] on the WIDER benchmark.

The second practical solution for detecting small objects is employing “context pooling” [6]. *CMS-RCNN* [27] pools

*Authors contributed equally
[1]The code is available at github.com/po0ya/face-magnet

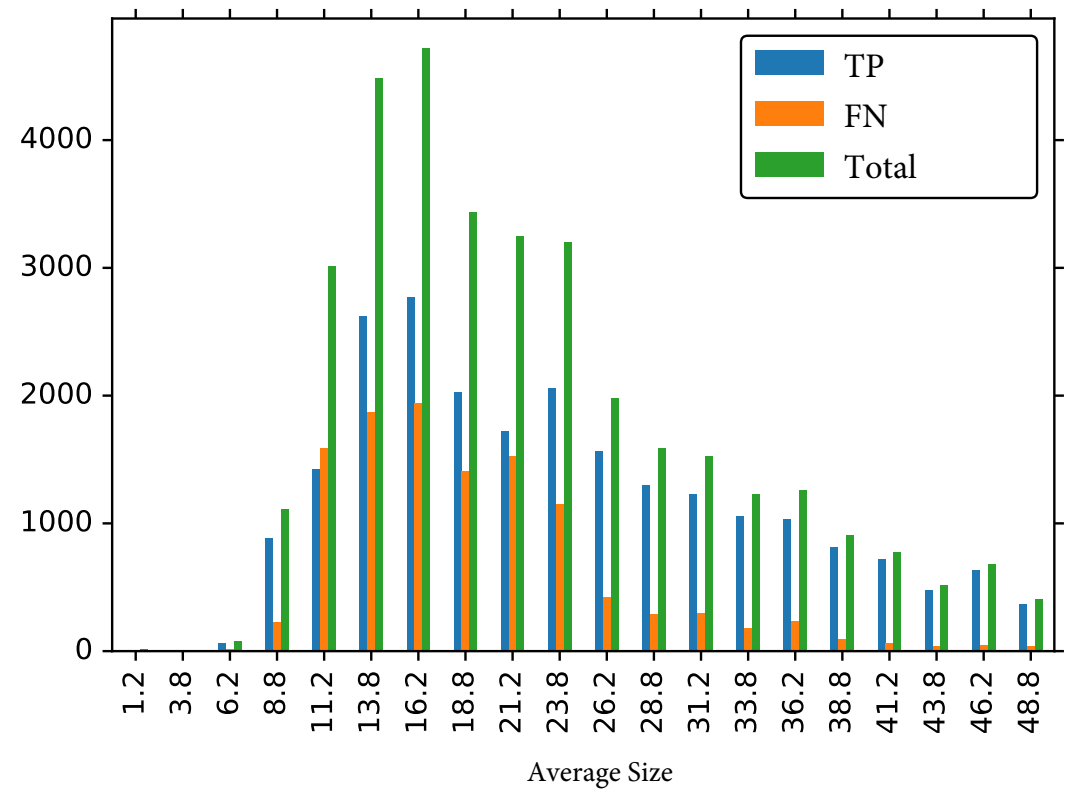

Figure 2: The distribution of the hits (True Positives) and misses (False Negatives) of a VGG16 network trained with the default settings of the Faster-RCNN framework between sizes 0 and 50 on the validation set of the WIDER dataset.

a larger box around the proposed boxes of a Faster-RCNN detector to account for the context information. We build on the same idea but we use a different network architecture and obtain better results than *CMS-RCNN* and HR-VGG16 with just using our context-augmented VGG16-based network.

Another way to tackle the scale challenge is to have multiple proposal and detection sub-networks. [2, 17] use this strategy to distribute object detection results at different scales to three sub-networks on a shared VGG16 core. We design the SizeSplit architecture which has two branches for small and big faces. Each branch has its own RPN and classifier. SizeSplit achieves comparable results to HR-VGG16 on the WIDER dataset.

Finally, we propose Face Magnifier Network (FaceMagNet) with the idea of discriminative magnification of feature maps before face proposal and classification. We show that this computationally efficient approach is as effective as size splitting and skip connections and when tested with an image pyramid beats the recently proposed ResNet101-based HR detector [7]. It is as efficient as a simple VGG16 based face detector which can detect faces with speed of 10 frames per second.

The rest of the paper is organized as follows. In Section 2, we review related works. Section 3 describes the baseline detectors. Section 4 presents our approach in detail. Ablation studies and benchmark results are presented in Section 5. Finally, we conclude the paper in Section 6.

## 2. Related Works

There is a vast body of research on face detection as it has been one of the oldest applications that machine learning proved to be practically useful via the Viola-Jones face detector [24]. In this section, we will only go over recent DCNN learning-based approaches. Most of the early DCNN-based face detectors are based on R-CNN[5] object detection which scores proposed regions by an external object proposal method. HyperFace [20] and DP2MFD [19] are some of the first face detectors based on this idea. After R-CNN, Fast-R-CNN [21] tried to remove the need for running the detector multiple times on object proposals by performing pooling on the feature maps. Faceness [22] improved the object proposals by re-ranking the boxes based on feature AP responses of the base network. The next development was removing the need for the external object proposal methods. Many methods such as [16] and Region Proposal Networks in the Faster-RCNN [21] were successful in eliminating the need for a pre-computed object proposals. Faster-RCNN trains the object proposal and classifier at the same time on the same base network. [10] is one of the earliest applications of Faster-RCNN in face detection. [27] as well as our work are also face detectors based on this framework.

There are also other detectors like those recently proposed in [17] and [7] that are object-proposal-less *i.e.* they perform detection in a fully convolutional manner as RPN does for proposing bounding boxes. Other methods like [11], [26] employ a cascade of classifiers to produce final detection results. Our final detector beats the cascade detectors and HR [7] while producing comparable results to SSH [17] on the WIDER face detection benchmark. However, these proposal-free methods are faster than Faster-RCNN-based methods since they are equivalent to the first part of such algorithms.

## 3. Baseline Detectors

The base object detector that we build on is Faster-RCNN [21]. It is a CNN that performs object detection in a single pass. Given an input image, it first generates a set of object proposals by its Region Proposal Network (RPN) branch. Then it does spatial max-pooling on the last convolutional layer feature maps to get the features for each proposed bounding box. The features are then used to classify and regress the dimensions of the candidate box. Our base architecture is a VGG16 network [23] with convolutional blocks *conv-1* to *conv-5* and fully connected layers $fc6$ and $fc7$. Each of the convolutional blocks are followed by a $2 \times 2$ max-pooling operator. Each block internally has two or three $3 \times 3$ convolutional layers of the same resolution. We also describe briefly the choice of hyper-parameters in the experiment section. We show the level of effectiveness of each design decision in Section 5.4.

### 3.1. Context Classifier

The final classifier can also make use of context information to detect smaller faces more accurately. We pool a con-text region around each candidate bounding box as shown in Figure 3 and concatenate the features to the pooled features of that bounding box. Then each are passed to a different set of fully connected layers, classifier and regressor. The final classification and regression is done on the concatenated embeddings of the context and the face branch. This is shown in Figure 3. We also tested another context model in which we merge the context and face features with a $1 \times 1$ convolution filter and then share the fully connected layers afterwards.

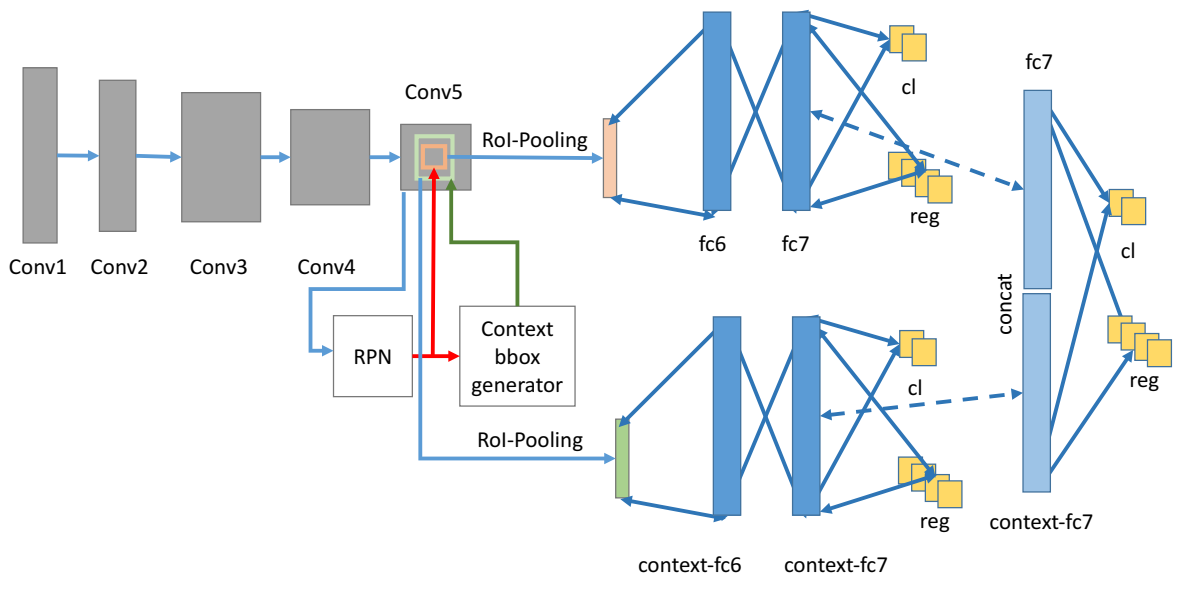

Figure 3: The architecture of the context network.

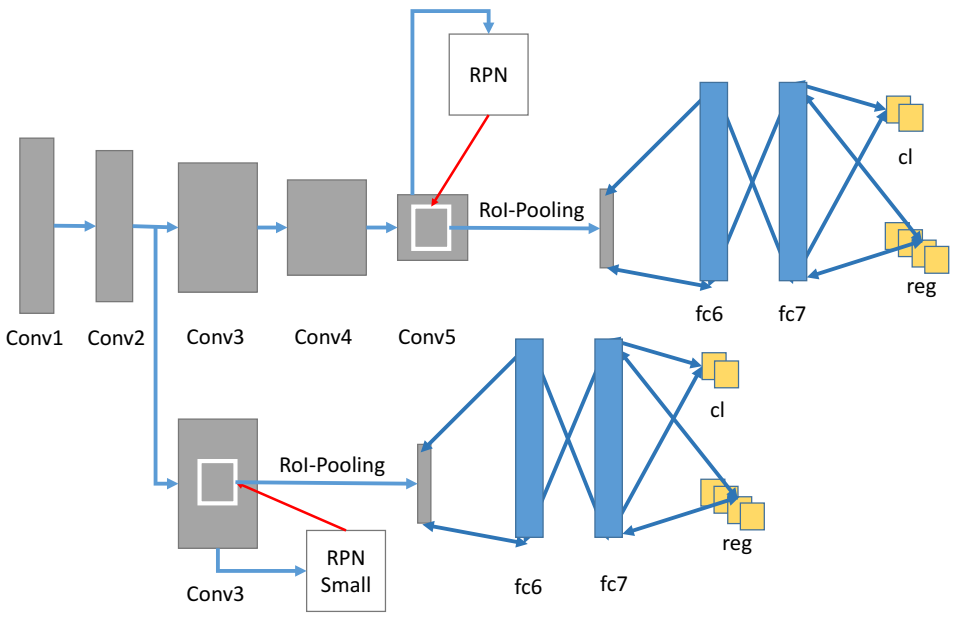

Figure 5: The architecture of the SizeSplit network.

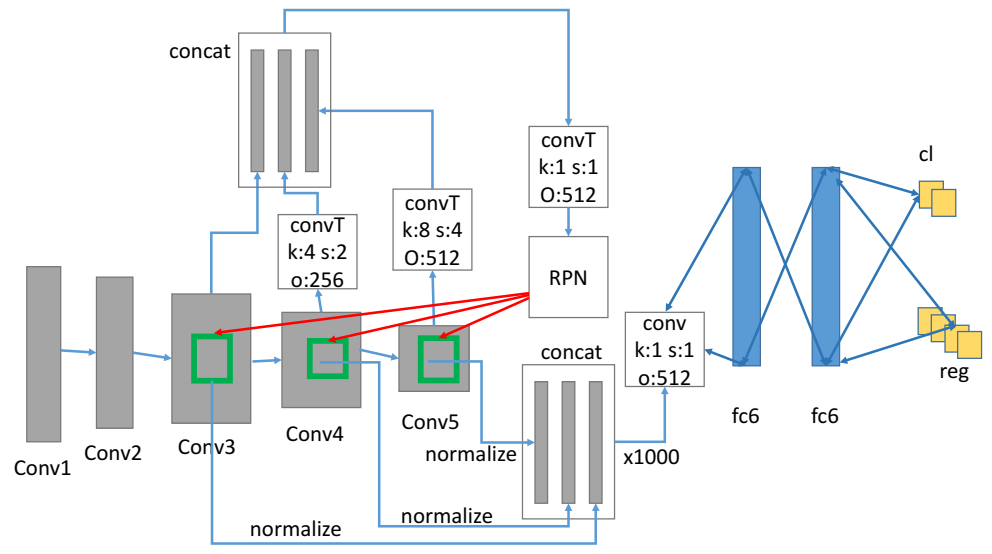

Figure 4: The architecture of the SkipFace network

### 3.2. Skip Pooling

Skip connections fuse information from shallower feature maps like conv-3 in VGG16, that have higher resolution with higher depth ones which are richer but have smaller dimensions. We fuse the APs by upsampling the deeper ones to the size of the biggest feature AP. One issue that arises in using skip connections is how to normalize the feature maps to be able to transfer the weights from a pre-trained base model effectively. [14, 27] normalize each feature column and then learns a separate weight per layer channel. Similar to [1], we do RoI-based block normalization. More precisely, let $\mathbf{X}_{conv-n} \in \mathcal{R}^{H \times W \times C}$ denotes the feature maps of a bounding box pooled from the convolutional block "n" where where $W$ and $H$ are the width and height of the pooled features of the $n$th block and $C$ is the number of channels. Then we define the normalized feature maps as $\mathbf{Y} = \mathbf{X} / \|\mathbf{X}\|_2$ and its derivative computed as

$$\frac{\partial \mathbf{Y}}{\partial \mathbf{X}} = \frac{\mathbf{I}}{\|\mathbf{X}\|_2} - \frac{\mathbf{X}\mathbf{X}^T}{\|\mathbf{X}\|_2^3} \tag{1}$$

is the globally normalized feature vector. The dimensions of these normalized features are then reduced using a 1x1 convolution layer appropriately to make transferring the weights from a pre-trained model possible. The architecutre of this method is shown in Figure 4. We employ the same strategy for the input of RPN network. Instead of normalizing over the pooled features, we normalize each block's final feature AP $\mathbf{X}^{rpn}_{conv-n} \in \mathcal{R}^{H \times W \times C}$ over the whole image, where $W$ and $H$ are the width and height of the final feature AP of *conv-n* respectively. Since $W$ and $H$ are different for each *conv-n* block, we employ up-sampling layers to rescale the normalized APs of *conv-4* and *conv-5* to the size of *conv-3*, increasing the receptive field for small faces. This was not needed for the classifier since $W$ and $H$ of the spatial pooling feature is the same for all layers. The upper part of Figure 4 shows our approach.

### 3.3. Size Specific Branches

Since features from lower depth layers are better for describing smaller objects, we design an architecture to accommodate this. It has a specialized branch for detecting small faces and another branch for detecting bigger ones. *conv-1* and *conv-2* of VGG16 are shared between the two branches. For big faces, we use the standard VGG16 architecture (*i.e.* blocks of *conv-3,4,5* and then *fc6,fc7*). To detect small faces, we have the *conv-3+conv-4-1* block and *fc6,fc7*. We use *conv-4-1* which has 512 output channels to be able to transfer the pre-trained weights since the input of *fc6* in the pre-trained Imagenet model has $7 \times 7 \times 512$ channels. The lower branch RPN and classifier predict boxes with one dimension less than 50 pixels, the upper branch detects faces with dimension greater than 50. An overview of this architecture is shown in Figure 5. Our network that implements this strategy beats HR-VGG16 [7] by 4% and *CMS-RCNN* [27] 14% on the validation set.

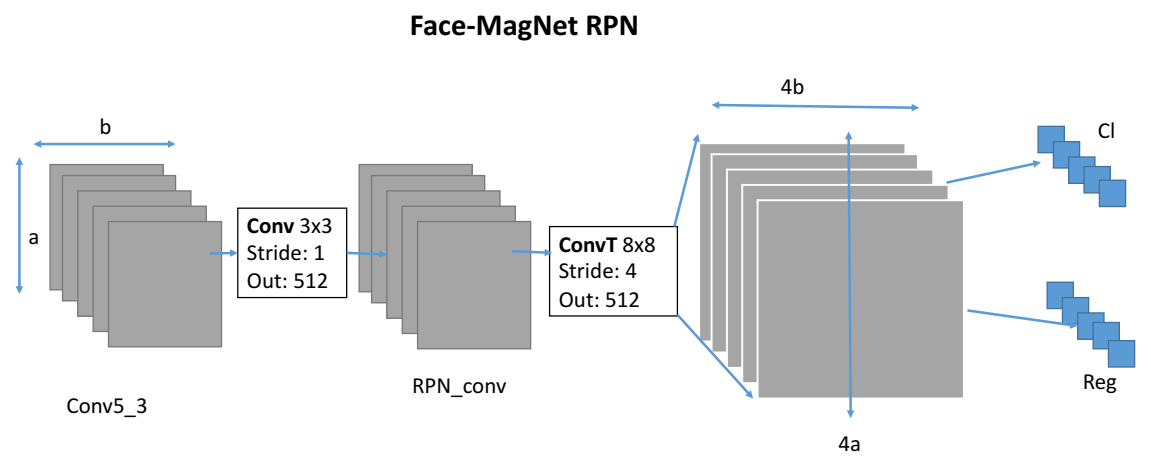

Figure 6: The proposed trainable RPN module to increase the resolution of object proposal module for a denser proposal.

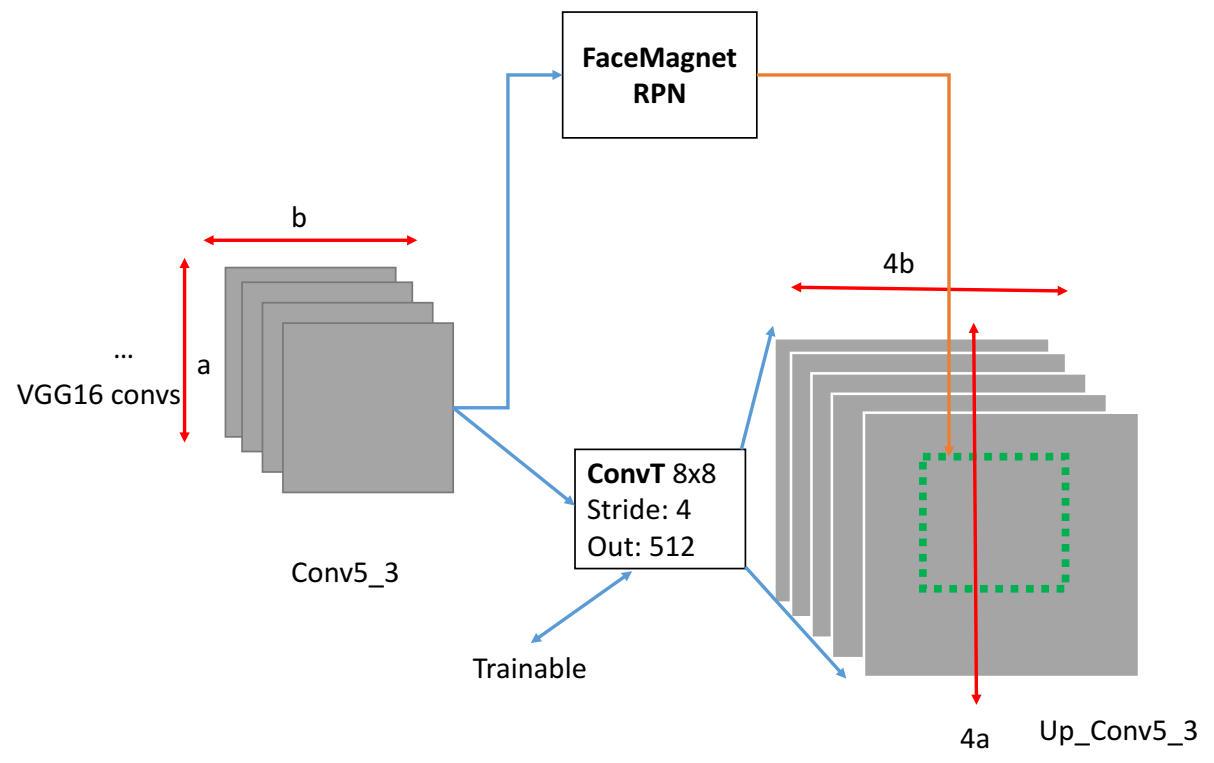

Figure 7: The architecture of the Face-MagNet network. Face-MagNet module can be found in Figure 6

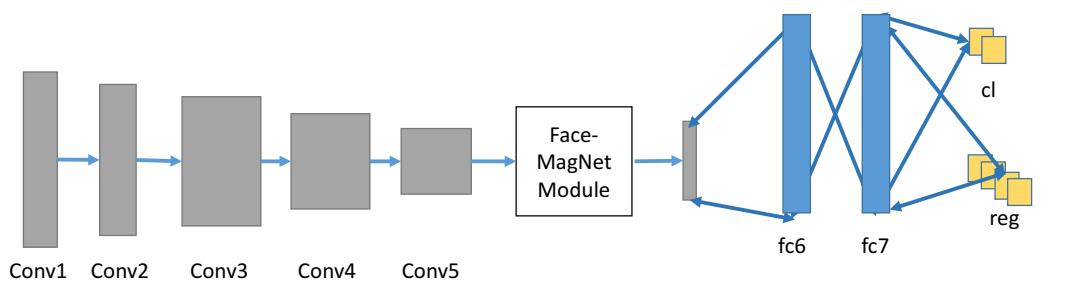

Figure 8: The architecture of the Face-MagNet network. Face-MagNet module can be found in Figure 7

# 4. Method

We propose to magnify the feature maps of the deeper layers in a discriminative way so that the feature AP resolution before the RPN, classifier, and the regressor increases. As a result, the RPN covers a denser grid on the original image which helps in proposing smaller bounding boxes. Moreover, the RoI pooling layer captures more information for smaller boxes. In the original design, $conv-5$ of VGG16 has a stride of 16 *i.e.* each location on the feature AP corresponds to 16 pixels in the original image. We train a $ConvTranspose$ layer, also known as deconvolution, before RoI pooling with kernel size of 8 and stride of 4. This way, each location will correspond to 4 pixels of the input image. We show that this proposed solution is almost as effective as adding context, skip pooling, and having multiple branches. The architecture of this approach is depicted in Figure8.

## 4.1. Loss

In this paper, all networks are trained to minimize a multi-task loss composed of classifier(s) and regressor(s) losses. In cases where there are more than one detection branch, the network is trained to minimize the loss of all branches simultaneously. The loss can be formulated as follows:

$$\mathcal{L} = \frac{1}{|\mathcal{S}|}\sum_{i\in\mathcal{S}}\ell_{cl}^{R}(A_i, G) + \frac{1}{|\mathcal{S}|}\sum_{i\in\mathcal{S}}\mathcal{I}(A_i \approx G) \otimes \ell_{R_r}(A_i, T_i)$$
$$+ \frac{1}{N}\sum_{i=1}^{N}\ell_{cl}^{C}(p_i, g_i) + \frac{1}{N}\sum_{i=1}^{N}\ell_{r}^{C}(p_i, t_i) \quad (2)$$

where the first two terms are the sum of the classification and regression losses of the RPN. $\mathcal{S}$ is the set of detection scales. $A_i$ is the set of anchors for the $i$'th scale. $G$ is the set of ground-truth annotations. $\mathcal{I}(A_i \approx G)$ is the indicator function and selects the anchors that have a nonzero overlap with a member of $G$. The second and third terms are the losses of the classifier and the regressor that score and regress the proposed boxes $p_i$. $g_i$ is the assigned ground-truth to bounding box $p_i$ and $N$ is the classifier batch size. Bounding boxes with an intersection over the union (IoU) of less than 0.3 are considered as negative examples and those with $IoU > 0.7$ are defined as positives. $l_{cl}^{*}$ is a binomial logistic loss and $l_{r}^{*}$ are the Robust-$L1$ losses defined in [4] and the supplementary material. $t_i$s are the regression targets which are defined the same way as [4] and described in the supplementary materials.

The total loss can be described as follows:

$$\mathcal{L} = \mathcal{L}_{rpn} + \mathcal{L}_{cl} \quad (3)$$

where $\mathcal{L}_{rpn}$ is the regression and classification loss of the RPN, and $\mathcal{L}_{cl}$ is the loss for the final regression and classification stage.

### 4.2. Context loss

In architectures with the context branch, the loss can be formulated as follows:

$$\mathcal{L}_{cl} = \alpha \mathcal{L}_{face} + \beta \mathcal{L}_{context} + \gamma \mathcal{L}_{joint} \tag{4}$$

where $\mathcal{L}_{face}$ is the loss on the *fc7* features, $\mathcal{L}_{context}$ is the loss on the *context-fc7* features, and $\mathcal{L}_{joint}$ is the loss on the concatenation of *fc7* and *context-fc7* features. Empirically, we have tried $\alpha = \beta = 0, \gamma = 1$,$\beta = 0, \alpha = \gamma = 1$ and $\alpha = \beta = \gamma = 1$ and the third set of values consistently performed better.

### 4.3. Train/Test Hyper-parameters

There are a number of training and testing phase hyper-parameters that can affect the face detection performance, especially for small faces. The scales of the RPN anchors are among the most important hyper-parameters. These scales should be chosen in a way that the proposed boxes cover the scale space of faces. [7] clusters the set of training boxes to define the anchor scales. We define these scales just to cover the minimum to maximum dimension of the boxes in the training data. Another important hyper-parameter is the input size during both training and inference. Whether to use image pyramids or not affects the detection performance, the extent of which depends on the base detector. The ratio of positive to negative examples at training time is usually set to $1 : 3$ in general object detection tasks, however we have consistently observed that a ratio of $1 : 1$ leads to a better face detection performance.

### 4.4. Final Networks

To show the effectiveness of our proposed Face Magnifier Network (Face-MagNet), we build two strong baseline classifiers: *SkipFace* and *SizeSplit*. These classifiers are trained and evaluated under the same settings and all of them surpass the performance of the recently proposed ResNet101-based HR method on the WIDER validation set. *SkipFace* represents the baseline designed based on the idea of the "skip connections" toward the scale challenge. The architecture consists of the Skip Pooling and the Skip RPN of 4, and the context classifier of 3. That is, the architecture is similar to Figure 4 with the context pooling added to the classifier from Figure 3. *SizeSplit* is another baseline that splits the predictions based on bounding box dimensions. It has one specialized branch for small and one branch for larger faces as shown in Figure 5. We add the context classifier to the lower branch which is responsible for detecting small faces. Therefore, the lower branch has the architecture of Figure 3. *Face-MagNet* represents the proposed Face Magnifier Network which has the base architecture discussed in 4 in combination with the context classifier of 3.1 to boost the performance on classifying smaller faces.

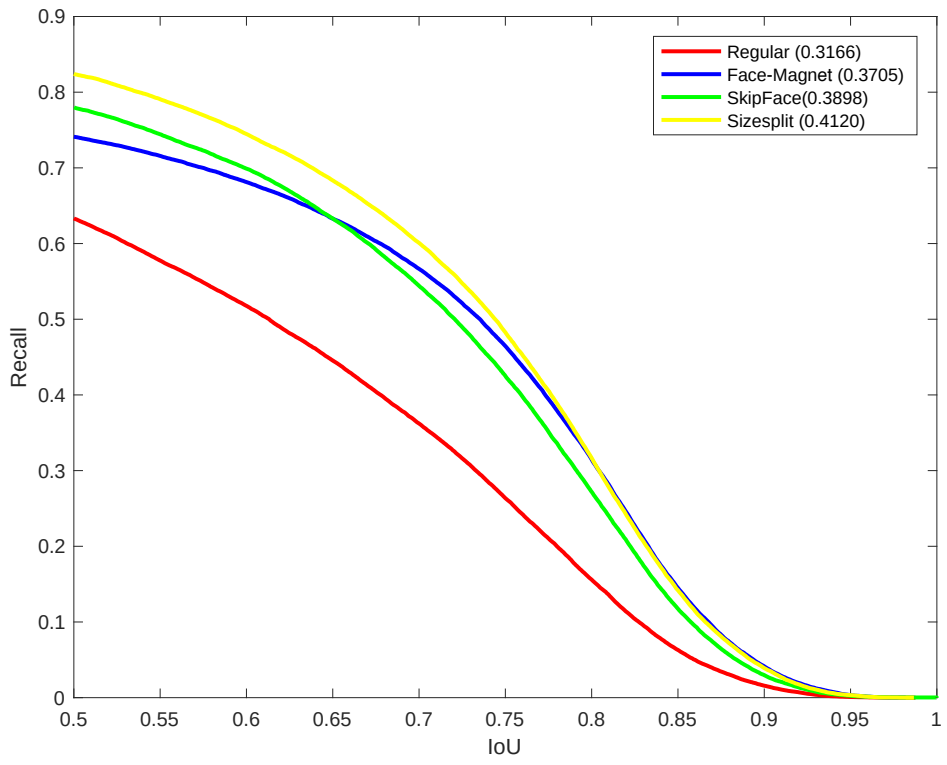

Figure 9: The IoU-Recall curve of the proposed and baseline models.

## 5. Experiments

In this section we train and evaluate models to study the design choices mentioned in the previous section to shed light on the relative performance of Face-MagNet in comparison to our strong baselines. In each study, the networks are trained and tested using the same hyper-parameters. All of the models are optimized with stochastic gradient descent algorithm with momentum for 38000 iterations. The initial learning rate is set to 0.001 and is reduced to 0.0001 after 30000 iterations. There are four images per mini-batch and the trainings are performed on two NVIDIA Quadro P6000 GPUs in parallel using Caffe deep learning framework [9]. The source code will be released publicly to ensure reproducibility.

### 5.1. Training Dataset

We have trained our networks on the training set of the WIDER [25] face detection dataset. There is a total of 32,203 images and with 393,703 annotated faces which have a high degree of variability in scale, pose and occlusion. 158,989 of the faces are for training, 39,496 are for validation, and the rest are for testing. Currently, it is the most challenging face detection dataset publicly available.

### 5.2. Localization Performance

As shown in the object detection benchmarks such as MS-COCO [13], the proposal stage is as important as the classification part for the detection systems since searching the space of all possible bounding boxes is not tractable. To compare the quality of the produced boxes explicitly, we plot the *"IoU-to-Recall"* curve of the proposed bounding boxes. This curve shows the recall versus the Intersection over Union (*IoU*) of the maximum overlapping bounding box. In other words, it represents the recall of the ground truth annotations (*i.e.* the percentage of the ground-truth an-notations covered with at least one box with an *IoU* greater than the specified threshold). Increasing this threshold decreases the number of covered annotations for all methods as expected. However, methods which have their curves fall on higher *IoU* thresholds are better as they have larger recalls at tougher localization criteria. Comparing recall values in general is not right because the precision is not considered. However, this comparison makes sense for RPNs because we are looking at the recall on top 1000 proposals and performing detection afterwards on these proposals.

Table 1: Table of Average Precision on hard partition of the validation set of WIDER. NC means without adding the context classifier module. Size means that the smallest dimension of the image is resized to that size keeping aspect ratio the same. We used scales of $[0.5, 1, 2]$ of the original size for the pyramid. The runtime column is for the minimum size of 1000 pixels and FDDB runtimes are average of runtime on original size inputs.

| Size | 600 | 800 | 1000 | 1200 | 1500 | Pyramid | Runtime | FDDB-Runtime |
|---|---|---|---|---|---|---|---|---|
| Base-VGG16 | 0.5002 | 0.6191 | 0.7094 | 0.7501 | 0.7629 | 0.7368 | 0.21s | 0.10s |
| SkipFace-NC | 0.7173 | 0.7657 | 0.7751 | 0.7734 | 0.773 | 0.7854 | 0.77s | 0.12s |
| SizeSplit-NC | 0.6479 | 0.7209 | 0.7577 | 0.7624 | 0.7624 | 0.7779 | 0.69s | 0.21s |
| Face-MagNet-NC | 0.6443 | 0.7353 | 0.7833 | 0.8003 | 0.7913 | 0.8170 | 0.77s | 0.10s |
| Context | 0.5983 | 0.7063 | 0.7658 | 0.786 | 0.794 | 0.7887 | 0.25s | 0.10s |
| SkipFace | 0.7453 | 0.7913 | 0.805 | 0.8064 | 0.806 | 0.8101 | 1.07s | 0.22s |
| SizeSplit | 0.6675 | 0.7367 | 0.768 | 0.7735 | 0.7733 | 0.8106 | 0.88s | 0.27s |
| Face-MagNet | 0.7006 | 0.7791 | 0.8156 | 0.82 | 0.8195 | 0.847 | 0.98s | 0.11s |
| CMS-RCNN [27] | 0.643 | - | - | - | - | - | - | - |
| HR-VGG16 [7] | - | - | - | - | - | 0.745 | - | - |
| HR-ResNet101 [7] | - | - | - | - | - | 0.806 | - | - |
| SSH [2] [17] | 0.686 | 0.784 | - | 0.814 | 0.810 | 0.845 | - | - |

Figure 9 shows this curve for the default input size configuration (*i.e.* minimum image side of 600 pixels and maximum of 1000 pixels). Figure 9 shows the localization results and their normalized area under the curve for the models. As can be seen at minimum size of 600 pixels, the localization performances of all three models are noticeably better than the default VGG16 base model. SizeSplit covers more bounding boxes since it has two RPN networks and produces twice as much of as the rest of the methods. The decay rate of the Face-MagNet is less than the other two approaches which shows it has more reliable proposals. Please note that the lower area under curve of the Face-MagNet for this image dimension compared to SizeSplit does not mean worse detection results as can be seen in the first column of Table 1.

### 5.3. Model Comparison and Input Size

To compare context pooling, skip connections, having specialized branches, and our feature AP magnification solution we train four VGG16-based face detectors with just that specific feature. We evaluate each model with different minimum input sizes and also image pyramiding. The training configuration for all of these models is the same and discussed in detail in supplementary materials. The upper part of Table 1 contains the results discussed in this subsection. As expected and shown in the first row in Table 1, all of the approaches improve the base VGG16 model. The first noticeable trend is that as the input size increases the performance on the hard partition of the WIDER benchmark gets imporved. Also, image Pyramid performs the best in all cases. SkipFace-NoContext which represents the skip connections solution without context pooling consistently outperforms SizeSplit-NC. SkipFace-NC and SizeSplit-NC do not benefit from increasing the input size in detecting smaller faces compared to the Context and Face-MagNet. This is consistent with the fact that earlier layer features are more sensitive to input distortions caused by up-sampling the input. Magnifying feature maps, or Face-MagNet architecture gives better results than the rest of the models. As it is just using deep features for classification so it is more robust to input size distortions. While at the same time, it has less complexity compared to the rest of the models. SizeSplit has the most number of parameters and than the Context model as it has another set of *fc6, fc7* fully connected layers for the context features.

### 5.4. Context

The rows corresponding to the SkipFace, the SizeSplit, and the Face-MagNet show the performance gain from adding context pooling to these base architectures. After the minimum size of 1000 pixels, our Face-MagNet architecture beats HR-ResNet101 and using pyramid pooling, it performs a bit better than the recent SSH face detector.

### 5.5. Image Pyramids

Image pyramids are conventional solution to detect faces with different scales. In Table 2, we show how different pyramiding strategies affect the performance of Face-

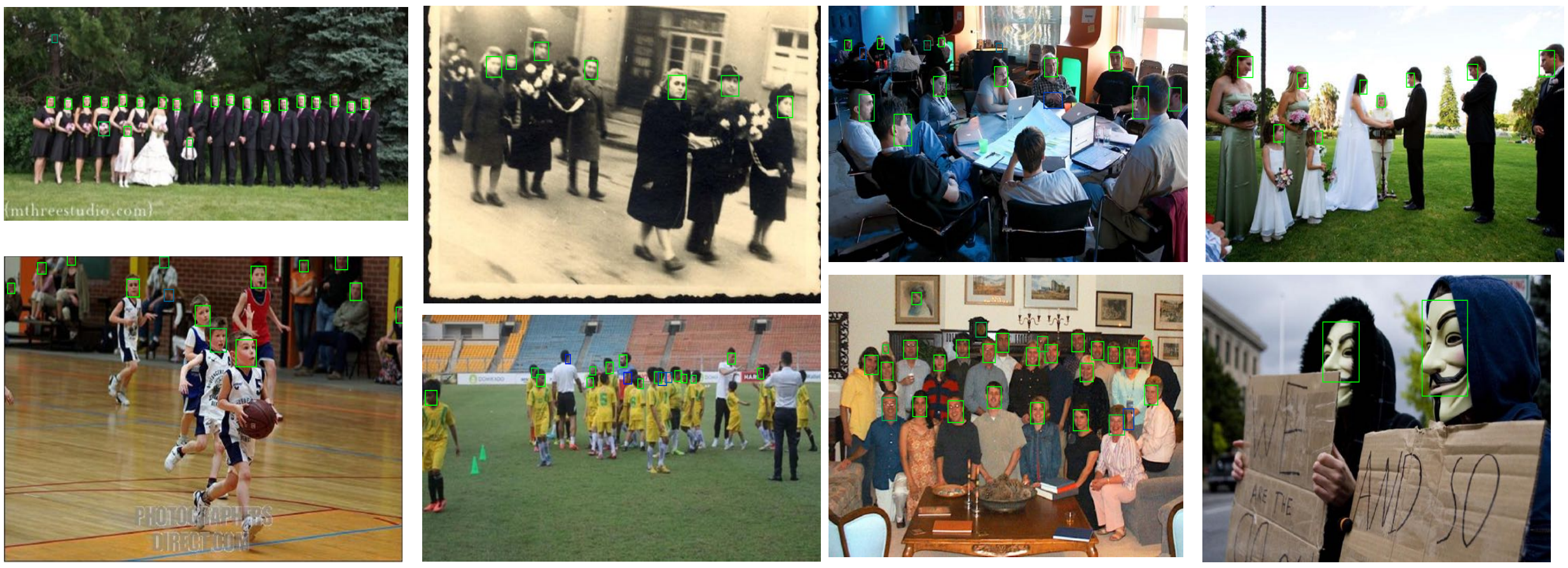

Figure 10: Qualitative results from WIDER dataset. The scores are reflected as the color of bounding boxes from Green to blue. Green corresponds to score 1.0 and blue to 0.5.

Table 2: Pyramid scale set effect on hard partition of WIDER validation set. Each scale set contains the up/down sampling ratio considered in the input image pyramid.

| Scale Set | 0.5,1,2 | 1,2 | 0.5,1,1.5,2 | 1,1.5,2 |
|---|---|---|---|---|
| Face-MagNet | 0.8423 | 0.8460 | 0.8470 | 0.8501 |

MagNet. The best pyramiding scales based on the validation set of WIDER is $\{1, 1.5, 2\}$.

### 5.6. Runtime Analysis

The runtime of the face detector depends on the input size of the networks. We test all the detectors on base size of 1000 pixels and also on FDDB dataset original input size. The experiments are done on a single nVidia Quadro P6000. The runtime of *Face-MagNet* on original size images is 0.11s which is almost the same as the *Base VGG16* model. It shows how efficient Face-MagNet is compared to the other models. On minimum size of 1000, the runtimes does not increase linearly since the GPU behavior depends on the architecture of the network. Face-MagNet magnifies the feature maps and thus when the input size increases the feature maps get larger too. It means that it takes up more memory than the original scale and also it gets slower since most of the time on the forward pass of the network is spent on the convolutional part of the network. The same trend happens for *SkipFace*, but not for the *Context* or *Base VGG16* model since the feature maps are smaller compared to the other models.

### 5.7. Detection Performance

We evaluate our approach on four face detection benchmarks. We use the same network that is trained on WIDER training set without any further fine-tuning. There are many

Table 3: The performance our final models on the validation set of the WIDER validation set. The bold values are the best performing in each column and the underline is for the second best.

| | Easy | Medium | Hard |
|---|---|---|---|
| Multiscale Cascade | 691 | 0.634 | 0.345 |
| Faceness-WIDER | 0.713 | 0.664 | 0.424 |
| Multitask Cascade | 0.848 | 0.825 | 0.598 |
| CMS-RCNN | 0.899 | 0.874 | 0.624 |
| HR-ResNet101 | 0.925 | 0.910 | 0.806 |
| SSH | **0.931** | **0.921** | 0.845 |
| SkipFace | 0.9228 | 0.9090 | 0.8101 |
| SizeSplit | 0.9266 | 0.9076 | 0.8106 |
| Face-MagNet | 0.920 | 0.913 | **0.850** |

results on these datasets; we select the top and recent ones for comparison. We compare our method with SSH[17], HR [7], Faster-RCNN [10, 21], Multiscale Cascade CNN [25], Faceness-WIDER [22], HyperFace [20], HeadHunter [15], LDCF [18], Conv3D [12]. Some qualitative detection results from all of the datasets are presented in Figure 10. We also report the results of the Face-MagNet on Pascal-Faces and FDDB datasets. Pascal-Faces contains annotated faces from Pascal VOC dataset which contains 851 images in unconstrained settings. The precision-recall curves for this dataset are shown in Figure 11. As can be seen, our method beats previously proposed face detection methods on this benchmark. The FDDB dataset [8] is another established dataset that is more challenging than PASCAL and AFW, since it has 2845 images with 5171 annotated faces. It has ten splits but, like most of the other methods; we evaluate our models on the whole dataset without resizing the input images with speed of 10 frames per second. As shown

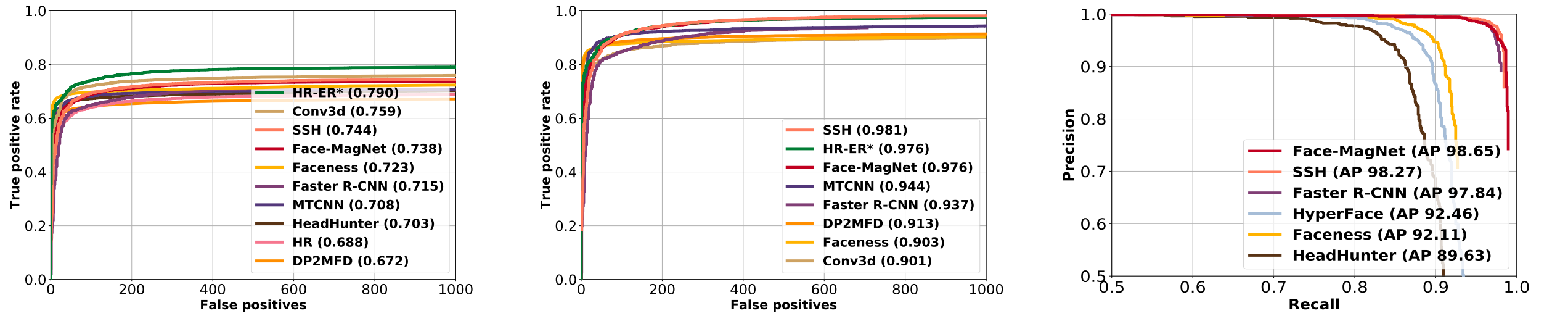

(a) FDDB continuous score. (b) FDDB discrete score. (c) Pascal-Faces.

Figure 11: Comparison among the methods on FDDB and Pascal-Faces datasets. (*Note that *HR-ER* is trained on the FDDB dataset in a 10-*Fold Cross Validation* fashion to do elliptical regression.)

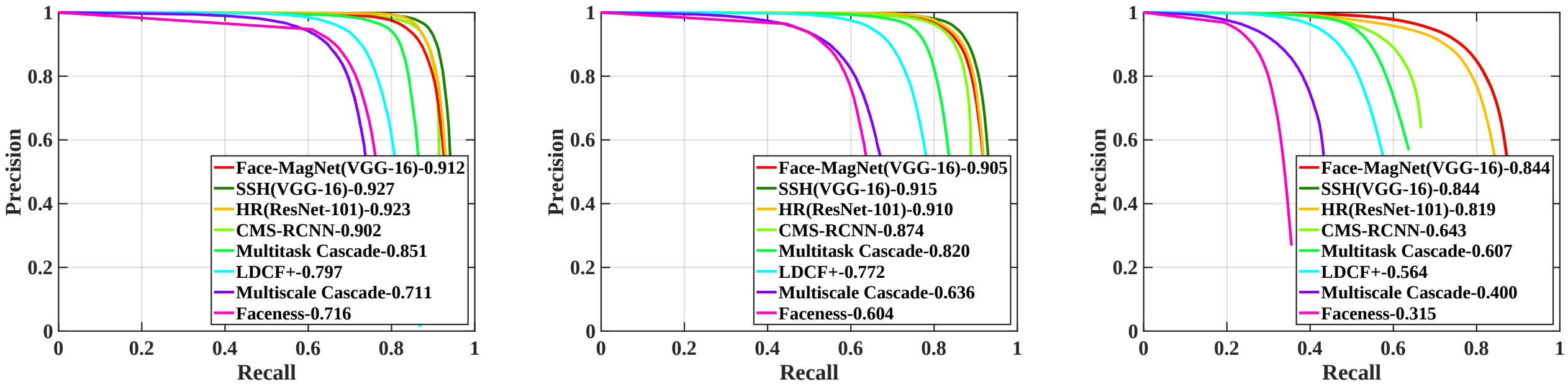

(a) Easy (b) Medium (c) Hard

Figure 12: Precision-Recall curves on WIDER test set.

in Figure 11 (a) and (b), we get close to state of the art on discrete and continouos scores.

Pascal Faces is annotated faces from Pascal VOC dataset which contains 851 images of unconstrained faces. The precision-recall curves for this dataset are shown in Figure 11 part (c). We test Face-MagNet on the original input size on this dataset. As can be seen, our method beats previously proposed face detection methods.

WIDER FACE [25] is the most challenging face detection benchmark with many small, occluded faces, various expressions, and poses. This dataset is split into three parts, easy, medium, and hard based on the difficulties regarding face scales. We present our results on this dataset in Figure 12 alongside few published methods on this recent benchmark following "scenario-int". The hard partition contains all the faces, including the ones in the easy and medium set of this benchmark. As it can be seen Face-MagNet outperforms the recently proposed methods.

## 6. Discussion

In this paper, we introduced Face-MagNet face detector that is capable of effectively detecting small faces. Face-MagNet employs $ConvTranspose$ layers inside RPN and classifier to magnify the feature maps for better detection of small faces. We developed three other architectures as our baselines for common solutions in the literature such as skip connections, context pooling, and detecting from different branches. We explored these solutions with extensive experiments and showed that the baselines in some settings outperforms HR-ResNet101. Finally, we showed that Face-MagNet outperforms the baselines and also recently proposed HR-ResNet101[7] significantly and achieves close results to the more recent SSH[17] face detector.

## Acknowledgement

This research is based upon work supported by the Office of the Director of National Intelligence (ODNI), Intelligence Advanced Research Projects Activity (IARPA), via IARPA R&D Contract No. 2014-14071600012. The views and conclusions contained herein are those of the authors and should not be interpreted as necessarily representing the official policies or endorsements, either expressed or implied, of the ODNI, IARPA, or the U.S. Government. The U.S. Government is authorized to reproduce and distribute reprints for Governmental purposes notwithstanding any copyright annotation thereon.

## References

[1] S. Bell, C. Lawrence Zitnick, K. Bala, and R. Girshick. Inside-outside net: Detecting objects in context with skip pooling and recurrent neural networks. In *Proceedings of the IEEE Conference on Computer Vision and Pattern Recognition*, pages 2874–2883, 2016.

[2] Z. Cai, Q. Fan, R. S. Feris, and N. Vasconcelos. A unified multi-scale deep convolutional neural network for fast object detection. In *European Conference on Computer Vision*, pages 354–370. Springer, 2016.

[3] M. Everingham, S. M. A. Eslami, L. Van Gool, C. K. I. Williams, J. Winn, and A. Zisserman. The pascal visual object classes challenge: A retrospective. *International Journal of Computer Vision*, 111(1):98–136, Jan. 2015.

[4] R. Girshick. Fast r-cnn. In *Proceedings of the IEEE International Conference on Computer Vision*, pages 1440–1448, 2015.

[5] R. Girshick, J. Donahue, T. Darrell, and J. Malik. Rich feature hierarchies for accurate object detection and semantic segmentation. In *Proceedings of the IEEE conference on computer vision and pattern recognition*, pages 580–587, 2014.

[6] B. Hariharan, P. Arbeláez, R. Girshick, and J. Malik. Hypercolumns for object segmentation and fine-grained localization. In *Proceedings of the IEEE Conference on Computer Vision and Pattern Recognition*, pages 447–456, 2015.

[7] P. Hu and D. Ramanan. Finding tiny faces. *arXiv preprint arXiv:1612.04402*, 2016.

[8] V. Jain and E. Learned-Miller. Fddb: A benchmark for face detection in unconstrained settings. Technical Report UM-CS-2010-009, University of Massachusetts, Amherst, 2010.

[9] Y. Jia, E. Shelhamer, J. Donahue, S. Karayev, J. Long, R. Girshick, S. Guadarrama, and T. Darrell. Caffe: Convolutional architecture for fast feature embedding. *arXiv preprint arXiv:1408.5093*, 2014.

[10] H. Jiang and E. Learned-Miller. Face detection with the faster r-cnn. *arXiv preprint arXiv:1606.03473*, 2016.

[11] H. Li, Z. Lin, X. Shen, J. Brandt, and G. Hua. A convolutional neural network cascade for face detection. In *Proceedings of the IEEE Conference on Computer Vision and Pattern Recognition*, pages 5325–5334, 2015.

[12] Y. Li, B. Sun, T. Wu, and Y. Wang. Face detection with end-to-end integration of a convnet and a 3d model. In *European Conference on Computer Vision*, pages 420–436. Springer, 2016.

[13] T.-Y. Lin, M. Maire, S. Belongie, J. Hays, P. Perona, D. Ramanan, P. Dollár, and C. L. Zitnick. Microsoft coco: Common objects in context. In *European Conference on Computer Vision*, pages 740–755. Springer, 2014.

[14] W. Liu, A. Rabinovich, and A. C. Berg. Parsenet: Looking wider to see better. *arXiv preprint arXiv:1506.04579*, 2015.

[15] M. Mathias, R. Benenson, M. Pedersoli, and L. Van Gool. Face detection without bells and whistles. In *European Conference on Computer Vision*, pages 720–735. Springer, 2014.

[16] M. Najibi, M. Rastegari, and L. S. Davis. G-cnn: an iterative grid based object detector. In *Proceedings of the IEEE Conference on Computer Vision and Pattern Recognition*, pages 2369–2377, 2016.

[17] M. Najibi, P. Samangouei, R. Chellappa, and L. Davis. Ssh: Single stage headless face detector. *Proceedings of International Conference on Computer Vision (ICCV)*, 2017.

[18] E. Ohn-Bar and M. M. Trivedi. To boost or not to boost? on the limits of boosted trees for object detection. *arXiv preprint arXiv:1701.01692*, 2017.

[19] R. Ranjan, V. M. Patel, and R. Chellappa. A deep pyramid deformable part model for face detection. In *Biometrics Theory, Applications and Systems (BTAS), 2015 IEEE 7th International Conference on*, pages 1–8. IEEE, 2015.

[20] R. Ranjan, V. M. Patel, and R. Chellappa. Hyperface: A deep multi-task learning framework for face detection, landmark localization, pose estimation, and gender recognition. *arXiv preprint arXiv:1603.01249*, 2016.

[21] S. Ren, K. He, R. Girshick, and J. Sun. Faster r-cnn: Towards real-time object detection with region proposal networks. In *Advances in neural information processing systems*, pages 91–99, 2015.

[22] C. C. L. Shuo Yang, Ping Luo and X. Tang. From facial parts responses to face detection: A deep learning approach. In *Proceedings of International Conference on Computer Vision (ICCV)*, 2015.

[23] K. Simonyan and A. Zisserman. Very deep convolutional networks for large-scale image recognition. *arXiv preprint arXiv:1409.1556*, 2014.

[24] P. Viola and M. J. Jones. Robust real-time face detection. *International Journal of Computer Vision*, 57(2):137–154, 2004.

[25] S. Yang, P. Luo, C. C. Loy, and X. Tang. Wider face: A face detection benchmark. In *IEEE Conference on Computer Vision and Pattern Recognition (CVPR)*, 2016.

[26] K. Zhang, Z. Zhang, Z. Li, and Y. Qiao. Joint face detection and alignment using multitask cascaded convolutional networks. *IEEE Signal Processing Letters*, 23(10):1499–1503, 2016.

[27] C. Zhu, Y. Zheng, K. Luu, and M. Savvides. Cms-rcnn: contextual multi-scale region-based cnn for unconstrained face detection. *arXiv preprint arXiv:1606.05413*, 2016.